\documentclass{article}

\PassOptionsToPackage{numbers, compress}{natbib}
\usepackage[final]{neurips_2020}
\usepackage[utf8]{inputenc} 
\usepackage[T1]{fontenc}    
\usepackage{hyperref}       
\usepackage{url}            
\usepackage{booktabs}       
\usepackage{amsfonts}       
\usepackage{nicefrac}       
\usepackage{microtype}      
\usepackage{algorithm}
\usepackage{algorithmic}
\usepackage{mathtools}

\usepackage{bm}

\usepackage{graphicx}
\usepackage{float}
\usepackage{hyperref}       
\usepackage{url}            
\usepackage{xcolor}         
\usepackage{subfigure}
\usepackage[graphicx]{realboxes}
\usepackage{amsmath}

\usepackage{enumitem}
\usepackage{pifont}         
\usepackage{multirow}       

\usepackage{graphicx}       
\usepackage{pythonhighlight} 
\usepackage{colortbl}

\title{FinGPT: Instruction Tuning Benchmark for Open-Source Large Language Models in \\ Financial Datasets}

\author{Neng Wang$^1$\thanks{Equal contribution.}, Hongyang (Bruce) Yang$^{2*}$, Christina Dan Wang$^3$\thanks{Corresponding author: Christina Dan Wang is Assistant Professor, Shanghai Frontiers Science Center of Artificial Intelligence and Deep Learning, NYU Shanghai; Business Division, NYU Shanghai, Shanghai China 200122. Email: christina.wang@nyu.edu. Supported in part by National Natural Science Foundation of China (NNSFC) grant 12271363},~\\ 
   $^1$University of California, Los Angeles;
   $^2$Columbia University; \\
   $^3$Shanghai Frontiers Science Center \\of Artificial Intelligence and Deep Learning, \\NYU Shanghai; Business Division, NYU Shanghai \\
  \texttt{nengwang19@ucla.edu; hy2500@columbia.edu; christina.wang@nyu.edu} \\  
       }

\begin{document}

\maketitle

\begin{abstract}

In the swiftly expanding domain of Natural Language Processing (NLP), the potential of GPT-based models for the financial sector is increasingly evident. However, the integration of these models with financial datasets presents challenges, notably in determining their adeptness and relevance. This paper introduces a distinctive approach anchored in the Instruction Tuning paradigm for open-source large language models, specifically adapted for financial contexts. Through this methodology, we capitalize on the interoperability of open-source models, ensuring a seamless and transparent integration. We begin by explaining the Instruction Tuning paradigm, highlighting its effectiveness for immediate integration. The paper presents a benchmarking scheme designed for end-to-end training and testing, employing a cost-effective progression. Firstly, we assess basic competencies and fundamental tasks, such as Named Entity Recognition (NER) and sentiment analysis to enhance specialization. Next, we delve into a comprehensive model, executing multi-task operations by amalgamating all instructional tunings to examine versatility. Finally, we explore the zero-shot capabilities by earmarking unseen tasks and incorporating novel datasets to understand adaptability in uncharted terrains. Such a paradigm fortifies the principles of openness and reproducibility, laying a robust foundation for future investigations in open-source financial large language models (FinLLMs). The codes have
been open-sourced at \url{https://github.com/AI4Finance-Foundation/FinGPT}.

\looseness=-1 
  
\end{abstract}

\section{Introduction}

Natural Language Processing (NLP) stands as a beacon for the financial sector \cite{xing2018natural,tai2013automatic,liu2022finrl}, offering groundbreaking opportunities and potential transformations. Large language models (LLMs) \cite{touvron2023llama1,brown2020language} grounded in the GPT framework are emerging as a focal point of interest, promising enhanced financial data interpretation and utilization. Instruction Tuning \cite{wei2021finetuned, ouyang2022training} efficiently adapts pre-trained LLMs to specific tasks, significantly saving time and computational resources without starting training from scratch. This cost-effective approach utilizes open-source models through an Instruction Tuning pipeline, achieving or even surpassing the performance of closed-source counterparts with minimal resource investment. Yet, the challenge remains in adeptly integrating these models, maintaining transparency, and ensuring their seamless adaptability to varied financial tasks.

Existing methods, while revolutionary in their own rights, fall short in certain key areas. Some models face obstacles in effortless integration with diverse financial datasets \cite{maia201818,malo2014good}. Others, although proficient in general contexts, might falter when exposed to intricate financial terminologies and scenarios \cite{araci2019finbert,wu2023bloomberggpt}. Hence, there's an evident need for a more comprehensive, transparent, and adaptable model catering specifically to the financial domain.


In this paper, we propose a novel scheme that utilizes the Instruction Tuning paradigm to magnify the capabilities of LLMs within the financial sphere. The proposed scheme stands out due to its emphasis on transparency, reproducibility, and the plug-and-play nature of model integration. Our works encompass the presentation of the Instruction Tuning paradigm, a deep-dive evaluation into the financial understanding of prevalent models, and an in-depth discussion of our methodological approach, which ensures a consistent training regimen. We bridge the gap between open-source models and financial data, ensuring a harmonious confluence.

Our contributions can be summarized as follows:

\begin{itemize}
\item \textbf{Instruction tuning paradigm:} We present an Instruction Tuning paradigm, specifically tailored for open-source Large Language Models (LLMs) in the financial sector. This approach not only addresses integration challenges but also enhances the adaptability and relevance of transformer-based models for various financial datasets.

\item \textbf{Cost-effective benchmarking scheme:} We introduce a benchmarking process designed with a cost-effective and end-to-end training and testing strategy. This scheme is not only tailored for financial contexts but also ensures a comprehensive and systematic evaluation of LLMs from basic competencies to complex, multi-task operations.

\item \textbf{Deep insights into various base models:} Our work offers a detailed exploration and clarification of various open-source base models such as Llama2, Falcon, ChatGLM2. By highlighting its potential for immediate and transparent integration into the financial sector, we provide valuable insights and plug-and-play guidance for researchers and practitioners working with financial tasks.


\item \textbf{Promotion of openness and reproducibility:} Our methodology adheres to and advocates for the principles of openness and reproducibility in the research and development of open-source FinLLMs. This contribution lays a solid foundation for future research, facilitating further investigation and development in the field.
\end{itemize}

The remaining sections of this paper are organized as follows: Section 2 shows the related work; Section 3 delves into the intricacies of the Instruction Tuning paradigm; Section 4 presents our implementation Detail; Section 5 outlines the experiment results; and Section 6 concludes the study with potential future directions.

\section{Related Works}

Recent years have seen a surge in research focused on amalgamating financial datasets with GPT-based models like GPT-3 and GPT-4 \cite{brown2020language} for enhanced NLP applications. Generally, there are two prevailing methodologies: Firstly, employing prompt engineering \cite{zhou2022large,white2023prompt,liu2022design} with open-source LLMs, keeping parameters intact; and secondly, using supervised fine-tuning methods such as Instruction Tuning \cite{ouyang2022training} to craft domain-centric LLMs specially designed for financial tasks.

\subsection{General Large Language Models}

\begin{itemize}
    \item \textbf{Llama2} \cite{touvron2023LLaMA2} is an open-source LLM developed by Meta, supports 20 languages, building upon its predecessor Llama 1 \cite{touvron2023llama1}. 

    \item \textbf{ChatGLM2} \cite{zeng2022glm,du2022glm} emerges as a bilingual model based on the General Language Model (GLM) framework \cite{DuQLDQY022}, supporting English and Chinese. 
    
    \item \textbf{BLOOM} \cite{scao2022bloom}, is the world’s largest open multilingual language model, supports 46 natural languages and 13 programming languages, serving as a comprehensive multilingual solution.
    
    \item \textbf{Falcon} \cite{falcon40b} is renowned for its multilingual support, efficiency in computation, and quality training data from diversified sources.
    
    \item \textbf{MPT} \cite{MosaicML2023Introducing} by MosaicML, pre-trained on English text and code, boasts an optimized architecture, making it efficient for both training and inference tasks.
    
    \item \textbf{Qwen} \cite{Qwen-VL} from Alibaba stands out for its prowess in both Chinese and English, making it a versatile tool for multilingual applications.
\end{itemize}

\subsection{Financial Large Language Models}
\begin{itemize}
\item \textbf{FinBert} \cite{araci2019finbert} is a dedicated model for financial sentiment analysis with under one billion parameters, fine-tuned on a rich financial corpus to excel in finance-specific tasks.

\item \textbf{FLUE} \cite{shah2022flue} offers a benchmark derived from five varied financial datasets, acting as an exhaustive evaluation tool for financial language understanding. Its derivative model, FLANG-BERT, outperforms FinBert on these datasets due to domain-specific enhancements.

\item \textbf{BloombergGPT} \cite{wu2023bloomberggpt} is a closed-source model based on BLOOM, trained extensively on diverse financial datasets, thereby encapsulating a broad spectrum of the financial domain.

    \item \textbf{FinGPT} \cite{yang2023fingpt,zhang2023instruct,fingpt_github} is an open-source LLM, fine-tuned from a general LLM using low-rank adaptation methods \cite{hu2021lora}, fostering accessibility for the broader community.
    \item \textbf{PIXIU} \cite{xie2023pixiu} functions as an evaluation benchmark and an instructional dataset. Its focus is solely on the dataset benchmark, exclusively evaluating models derived from Llama without considering other open-source LLMs.

\end{itemize}


Current research mainly uses Llama models as the base model for financial task evaluations, limiting understanding as different open-source models may excel in various tasks. A broader, more inclusive evaluation encompassing various open-source models could yield insights into task-specific performances and may unveil models that are inherently better aligned with specific financial applications. 

Acknowledging these gaps, our work endeavors to present a more sophisticated and integrated paradigm, aiming to seamlessly intertwine open-source LLMs with intricate financial data, thereby proposing a robust solution for domain-specific applications.


\section{Proposed Paradigm}
\begin{figure}
\centering
\includegraphics[scale=0.38]{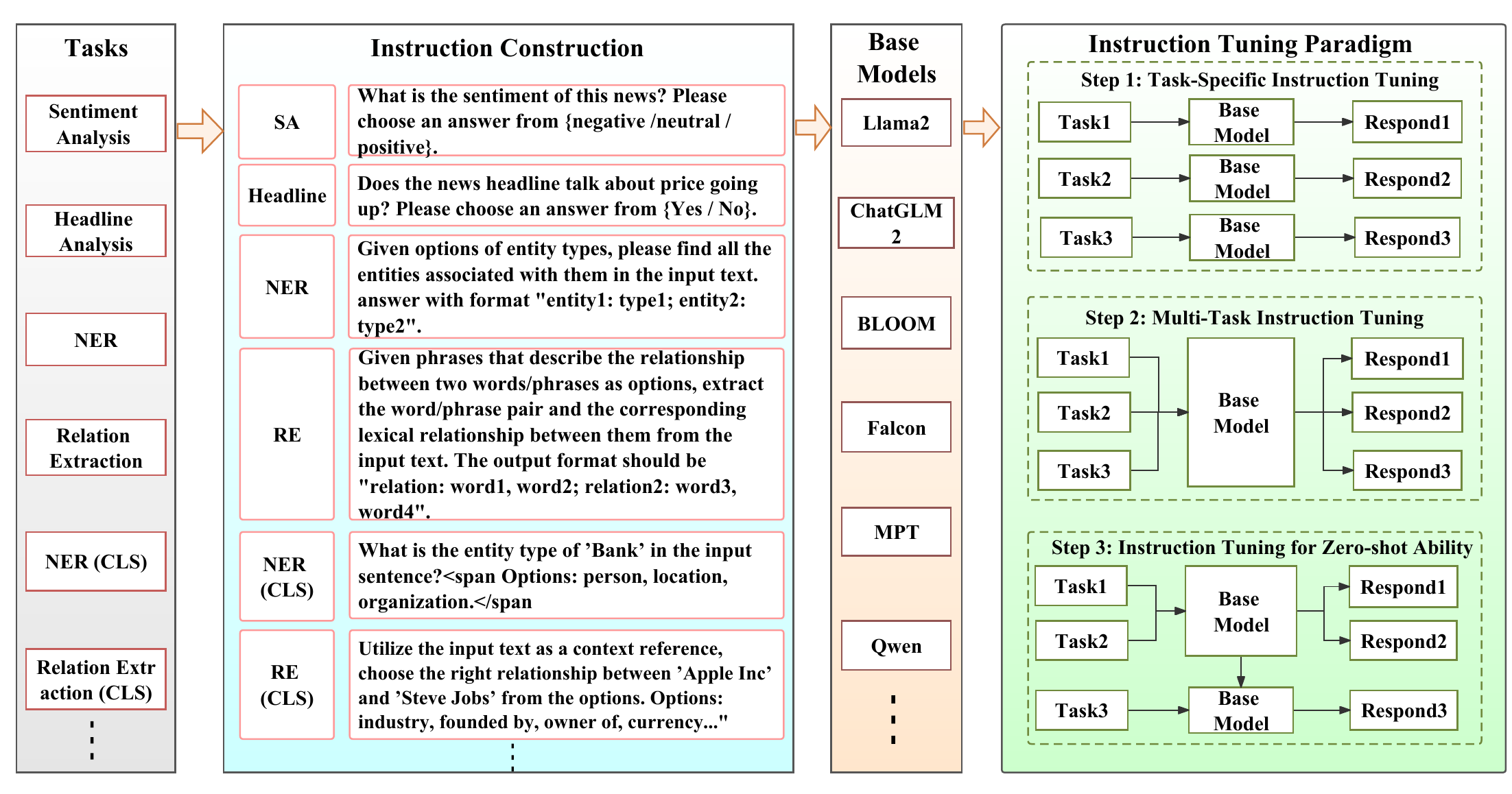}
\caption{Overview of the proposed Instruction Tuning paradigm}
\label{fig:framework}
\end{figure}

Initially, we conduct a continuous evaluation of financial tasks while constructing instructions pertinent to each task, followed by the selection and evaluation of a base model. The Instruction Tuning paradigm outlined in this study, depicted in Figure \ref{fig:framework}, is carefully designed into three interconnected phases, each playing a crucial role in facilitating the seamless integration and thorough analysis of various financial NLP datasets.

\subsection{Task-Specific Instruction Tuning}
In the initial phase of our paradigm, Task-Specific Instruction Tuning, we meticulously analyze the foundational competencies of LLMs for individual NLP tasks within the finance sector. Each task is examined in isolation during this phase, allowing for a detailed evaluation of LLMs’ inherent capabilities and performance. This focused approach generates in-depth insights into the strengths, efficacy, and areas needing improvement for each LLM regarding specific tasks and the ability to efficiently extract, process, and analyze information from various financial data sources.

\subsubsection{Potential Challenges in Task-Specific Instruction Tuning}
This dedicated approach, while robust, is not without challenges:

\begin{itemize}    
    \item \textbf{Varying Task Complexity:} Financial NLP tasks vary considerably in their level of complexity and specificity. As such, LLMs might exhibit proficiency in certain tasks while struggling with others that demand a deeper understanding of the financial domain or more advanced analytical skills.
        
    \item \textbf{Quality of Financial Datasets:} The quality and reliability of financial datasets used for task-specific instruction tuning directly impact the effectiveness of the tuning process. Ensuring data accuracy, relevance, and completeness while avoiding biased or unrepresentative samples is crucial for the success of this phase.
    
    \item \textbf{Performance Measurement:} Establishing appropriate metrics and benchmarks to accurately measure and compare the performance of LLMs across various task-specific scenarios can be challenging given the unique characteristics of each task within the financial domain.
\end{itemize}

\subsubsection{Addressing the Challenges in Task-Specific Instruction Tuning}

Addressing the outlined challenges requires a multifaceted strategy combining data quality assurance, enhanced model training and testing procedures, and continuous performance monitoring.

\begin{enumerate}    
    \item Recognizing the varied complexity levels across different financial NLP tasks, we implement a dynamic Instruction Tuning approach. This approach is adaptive, changing the depth and breadth of tuning based on the complexity of the task at hand, thus ensuring optimal performance across tasks of varying difficulty and specificity.
    
    \item Rigorous data validation and verification processes are instituted to ensure the quality and reliability of financial datasets utilized in the tuning process. These processes aim to verify data accuracy, completeness, and relevance, thus providing a solid foundation for effective task-specific Instruction Tuning.
    
    \item We devise a set of comprehensive, task-appropriate performance metrics and benchmarks. Continuous refinement and validation of these metrics ensure they remain relevant and reflective of the unique characteristics and requirements of each task.
\end{enumerate}

By addressing these challenges head-on, we ensure that the task-specific phase of our paradigm is both robust and reflective of the unique demands and challenges posed by the financial domain.

\subsection{Multi-Task Instruction Tuning}

The Multi-Task Instruction Tuning phase evaluates the LLMs' versatility and adaptability across concurrent NLP tasks within finance. Here, various instructional tunings are integrated, enabling LLMs to perform multiple tasks simultaneously, offering insight into their multitasking capabilities. 

This phase is designed to mirror the complex, multitasking environment that characterizes the financial sector, where the ability to concurrently process and analyze various forms of data is paramount. Therefore, it critically informs the practical utility and efficiency of deploying LLMs in real-world financial scenarios.

\subsubsection{Potential Challenges in Multi-Task Instruction Tuning}
While this phase is imperative, it introduces several challenges that must be navigatively addressed:

\begin{itemize}
    \item \textbf{Task Interference:} One major challenge is task interference. When models are trained to perform multiple tasks concurrently, the learning for one task might interfere with the learning for another, affecting the overall performance adversely.
    
    \item \textbf{Computational Complexity:} With the amalgamation of instructional tunings for various tasks, the computational complexity increases. Handling the elevated processing demands without compromising on efficiency and speed becomes challenging.
    
    \item \textbf{Optimal Task Weighting:} Determining the appropriate balance or weighting among multiple tasks during training to ensure that no single task dominates the learning process is a non-trivial challenge.
\end{itemize}

\subsubsection{Addressing Challenges in Multi-Task Instruction Tuning}

Addressing the inherent challenges in Multi-Task Instruction Tuning requires a streamlined approach combining optimized computation, efficient training protocols, and advanced evaluation techniques.

\begin{enumerate}
    
    \item Advanced optimization techniques are employed to handle increased computational demands. Through efficient batching and parallel processing, computational loads are effectively distributed, ensuring fast and efficient training without sacrificing model quality.
    
    \item Dynamic task weighting strategies are adopted to facilitate balanced learning across tasks. These adaptive mechanisms adjust the weight assigned to each task’s loss during training, promoting harmonious multi-task learning without dominance of any single task.
\end{enumerate}

These strategic measures collectively address the challenges associated with Multi-Task Instruction Tuning, enabling the effective deployment of proficient LLMs in the financial domain.

\subsection{Instruction Tuning for Zero-shot Ability}

The final phase, “Instruction Tuning for Zero-shot Ability,” enhances LLMs' zero-shot capabilities. In this crucial phase, LLMs face unprecedented scenarios and tasks, selected to assess their adaptability, learning agility, and response to novel challenges in the financial sector. The introduction of novel datasets and unseen tasks creates a rigorous testing environment for the models. This systematic approach allows for a detailed examination of the LLMs' robustness, flexibility, and problem-solving abilities, essential for operating in the rapidly changing financial landscape.

\subsubsection{Potential Challenges in Instruction Tuning for Zero-shot Ability}

The complexity and novelty integrated into this phase inevitably introduce an array of challenges, each of which necessitates thoughtful consideration and strategic addressing:

\begin{itemize}
    \item \textbf{Hallucination:} LLMs sometimes generate plausible but unfounded or hallucinated information in their responses, particularly when dealing with unfamiliar or unseen tasks. This phenomenon can lead to misinformation and misinterpretation of the data, which is especially perilous in the financial domain where precision is paramount. 
        
    \item \textbf{Generalization vs. Specialization:} Striking the optimal balance between generalization to new tasks and specialization in previously learned tasks is a perpetual challenge in zero-shot learning environments.
\end{itemize}

\subsubsection{Addressing Challenges in Instruction Tuning for Zero-shot Ability}

To effectively navigate through the challenges identified, strategic approaches have been employed:

\begin{enumerate}
    \item To counteract hallucination issues, we implement a strategy of task reformulation. The models are trained with a more focused objective, making it easier for them to understand and categorize input data without generating extraneous information. This focused training approach narrows down the task gap between the newly reformulated tasks and the target task, thereby fostering a more controlled and accurate generation process.


    \item An optimal balance between generalization to novel tasks and specialization in learned tasks is crucial. We use adaptive learning and fine-tuning techniques, dynamically adjusting the learning process based on the task at hand. 
\end{enumerate}

Through these carefully devised strategies, we mitigate identified challenges, promoting the effective development and tuning of LLMs for enhanced zero-shot capabilities in the financial domain.

\section{Implementation Detail}

In this section, we expound upon our methodologies in Data Preparation, Instruction Construction, and Training. Comprehensive resources including datasets, codebases, and illustrative examples are accessible at \url{https://github.com/AI4Finance-Foundation/FinGPT/tree/master/fingpt/FinGPT_Benchmark}.

\subsection{Data Preparation}

In preparing the data, our approach aligns with the methodology utilized by BloombergGPT\cite{wu2023bloomberggpt}, in which a selection of financial datasets from the FLUE benchmark\cite{shah2022flue} is adopted for various tasks.

\textbf{Selection of Datasets:} For the Sentiment Analysis (SA) task, we leverage datasets FPB\cite{malo2014good} and FiQA-SA\cite{maia201818}, while for the Headline Classification (HC) task, the Headline dataset\cite{sinha2020impact} is employed. Furthermore, the NER dataset\cite{salinas-alvarado-etal-2015-domain} is utilized for Named Entity Recognition (NER) tasks. To enhance diversity within the data, particularly for the sentiment analysis task, additional datasets, TFNS\cite{tfns2022} and NWGI\cite{fingpt_github}, were incorporated into the mix. This step was crucial to ensure that the models developed had exposure to a wide variety of data, promoting robustness and versatility in their application.

\textbf{Financial Relation Extraction:} Additionally, our implementation ventured into the domain of financial Relation Extraction (RE). For this endeavor, we engaged with the FinRED dataset\cite{10.1145/3487553.3524637}, providing a basis for the extraction and understanding of financial relations within the textual data.

\textbf{Rationale Behind Dataset Choices:} Datasets were carefully selected to cover a wide range of financial NLP tasks, with each contributing uniquely to the model's understanding and performance on specific tasks. Care was taken to ensure that the datasets were complementary, and their integration would facilitate a comprehensive and nuanced understanding of the model's capabilities and areas that required further refinement and tuning.

\renewcommand{\arraystretch}{1.3}
\begin{table}[h]
\centering
\begin{tabular}{|c|c|c|c|c|c|c|}
\hline
Task & Dataset & Total Samples \\
\hline
\multirow{4}{*}{Sentiment Analysis (CLS)} & FPB & 3634 \\
\cline{2-3}
& FiQA-SA & 938 \\
\cline{2-3}
& TFNS & 9543 \\
\cline{2-3}
& NWGI & 16184 \\
\hline
Named Entity Recognition & NER & 609 \\
\hline
Headline Classification (CLS) & Headline & $11412 \times 9$ \\
\hline
Relation Extraction & FinRED & 6768 \\
\hline
Named Entity Recognition (CLS) & NER & 1003 \\
\hline
Relation Extraction (CLS) & FinRED & 9657 \\
\hline
\end{tabular}
\vspace{2mm}
\caption{Overview of tasks and datasets: The trailing (CLS) means the task can be formatted into a unified classification task. Number of samples are counted before augmented by hand-craft/gpt-generated prompts. Headline contains 11412 sample and 9 question for each sample. For each entity in NER, we ask models to classify its entity-type forming NER(CLS). For each relation in RE, we ask models to classify its relation-type forming RE(CLS).}
\label{table:datasets}
\vspace{-2mm}
\end{table}
\renewcommand{\arraystretch}{1.0}

\subsection{Instruction Construction}

Constructing precise instructions is pivotal for both task-specific and multi-task Instruction Tuning, with each task being guided by a unique instruction prompt.

\textbf{Template Structure:} The instruction template is structured as follows: 

\textbf{Instruction}: [prompt] \textbf{Input}: [input] \textbf{Answer}: [output]

This template provides a standardized format, facilitating consistency across different tasks and experimental setups.

\textbf{Instruction Formulation for Specific Tasks:}
\begin{itemize}
    \item \textbf{Sentiment Analysis (SA) Task:} Instructions for the SA task are directly adopted from FinGPT \cite{zhang2023instruct}, leveraging their previously established efficacy.
    \item \textbf{Headline Classification (HC) Task:} For the HC task, we have chosen to use Bloomberg’s \cite{wu2023bloomberggpt} set of instructions, taking advantage of their industry-aligned approach.
    \item \textbf{Named Entity Recognition (NER) and Relation Extraction (RE) Tasks:} In the cases of NER and RE, instructions were meticulously crafted in-house to address the specific nuances and requirements of these tasks.
\end{itemize}

\textbf{Multi-Task and Zero-Shot Experiment Adjustments:} To facilitate multi-task and zero-shot experiments, modifications were made to the NER and RE tasks, reformatting them into classification tasks—dubbed NER(CLS) and RE(CLS). This alignment with SA and HC tasks not only enriched our set of tasks but also aimed to enhance the generalization capabilities of the LLMs. 

\textbf{Zero-Shot Experiment Instructions:} An [Options] section is added to each instruction for standardization in zero-shot experiments, streamlining instructions and limiting unexpected answers. ChatGPT was used to create ten unique instructions per task, with option order randomized in the training set for diversity. The updated zero-shot experiment template is provided below:

\textbf{Instruction}: [prompt] \textbf{Options}: [options] \textbf{Input}: [input] \textbf{Answer}: [output]

Concrete examples illustrating the constructed instructions are depicted in Figure \ref{fig:framework}, providing visual insights into the practical application of the instruction set within the experimental framework.





\subsection{Training Detail}

In our research, six open-source LLMs—Llama2-7B\cite{Llama2_7B}, Falcon-7B\cite{Falcon_7b}, BLOOM-7.1B\cite{BLOOM_7b1}, MPT-7B\cite{MPT_7B}, ChatGLM2-6B\cite{ChatGLM2_6B}, and Qwen-7B\cite{Qwen_7B}—are selected as the subjects for the application of our Instruction Tuning paradigm. Each of these models is of a comparable size and is used in their base form, without the inclusion of any instruction-tuned variants or chat versions, with the singular exception of ChatGLM2 (the base model of which has not been released).

\textbf{Utilizing LoRA.}
Given the substantial computational resources necessitated by the fine-tuning process of these LLMs, our approach incorporates the use of LoRA\cite{hu2021lora}, maintained with a rank of 8 and a scaling factor (alpha) of 32, targeting the projection layers within attention modules.

In the three phases of Instruction Tuning:
\begin{itemize}
    \item \textbf{Task-specific job}: epochs are set to 8 for SA, HC, and RE tasks due to their ample sample sizes. The NER task, possessing a limited sample size, warrants the setting of epochs to 50.
    \item \textbf{Multi-task job}: the models are exposed to a combined dataset of SA, HC, NER, RE, NER(CLS), and RE(CLS) for a total of 4 epochs. Tasks characterized by smaller sample sizes are oversampled to maintain balance.
    \item \textbf{Zero-shot job}: the models undergo fine-tuning on a consolidated dataset comprised of NER(CLS), RE(CLS), and HC for a single epoch, subsequently undergoing evaluation on the SA task. Checkpoints are systematically saved every 100 steps and are selected based on their evaluation loss pertaining to the SA task.
\end{itemize}


\textbf{Model Parameters:}
Experiments used four RTX 3090 GPUs, with max token length of 512 and per-device batch size of four, plus eight gradient accumulation steps. We employed AdamW optimizer\cite{loshchilov2019decoupled}, with an initial learning rate of \(1 \times 10^{-4}\), linearly decaying to zero following a 3\% warm-up in steps, utilizing FP16 precision for cost-effectiveness.


\textbf{Training Cost Analysis:}
With GPU hourly rate at \$3.36, task-specific jobs for six base models across four tasks took 30 hours. Multi-task and zero-shot jobs required about 60 hours due to increased instructions per base model, totaling 90 hours. Thus, the entire training cost was \$302.4.

\renewcommand{\arraystretch}{1.3}
\begin{table}[h]
\centering
\begin{tabular}{|c|c|c|c|c|c|c|}
\hline
Dataset & Llama2 & Falcon & MPT & BLOOM & ChatGLM2 & Qwen \\
\hline
SA & 0.820 (2) & 0.804 (4) & \textbf{0.821} (1) & 0.748 (6) & 0.798 (5) & 0.811 (3) \\
NER & 0.673 (3) & 0.619 (5) & 0.615 (6) & \textbf{0.729} (1) & 0.645 (4) & 0.679 (2) \\
HC & \textbf{0.942} (1) & 0.940 (3) & 0.938 (4) & 0.930 (6) & \textbf{0.942} (1) & 0.936 (5) \\
RE & 0.395 (3) & \textbf{0.428} (1) & 0.309 (6) & 0.425 (2) & 0.340 (5) & 0.371 (4) \\
\hline
Avg Ranking & \textbf{2.0} & 3.25 & 4.25 & 3.75 & 3.75 & 3.5 \\
\hline
\end{tabular}
\vspace{2mm}
\caption{Task-Specific Instruction Tuning Results Summary: Each row presents the F1-score of base models tuned on the specified task, along with their rankings in the bracket. The best model in each task is highlighted in bold. The average of rankings is computed to assess overall performance.}
\label{table:phase1_all}
\vspace{-2mm}
\end{table}
\renewcommand{\arraystretch}{1.0}

\section{Experiment Results}

\subsection{Task-Specific Instruction Tuning}

\textbf{Result Overview.}
Table \ref{table:phase1_all} summarizes the results of task-specific Instruction Tuning. For Sentiment Analysis (SA), we document the average performance of models across all sub-datasets, represented through the mean F1-score, with detailed performance metrics for each SA dataset available in Table \ref{table:phase12_sa}. The entity-level F1-score is reported for Named Entity Recognition (NER), whereas for Health Classification (HC) and Relation Extraction (RE), the F1-scores are reported respectively. For RE, we solely consider the F1-score for identified relations, excluding the (relation, subject, object) tuples.

\textbf{Performance Insights.}
The experimental findings yield interesting insights. Notably, Llama2 delivers superior overall performance, as evidenced by its average ranking of second place across all tasks. Both Falcon and Qwen demonstrate versatility, yielding balanced performances across all tasks under consideration. In contrast, while BLOOM excels significantly at Information Extraction (IE) tasks such as NER and RE, it falls short in classification tasks like SA and HC, where it records the lowest performance. Although MPT secures the highest score in SA, it underperforms in NER and RE tasks.

\renewcommand{\arraystretch}{1.3}

\begin{table}[h]
\centering
\begin{tabular}{|c|c|c|c|c|c|c|c|}
\hline
Phase & Dataset & Llama2 & Falcon & MPT & BLOOM & ChatGLM2 & Qwen \\
\hline
\multirow{5}{*}{Task-Specific} & FPB & 0.863 & 0.846 & \textbf{0.872} & 0.810 & 0.850 & 0.854 \\
& FiQA & \textbf{0.871} & 0.840 & 0.863 & 0.771 & 0.864 & 0.867 \\
& TFNS & 0.896 & 0.893 & \textbf{0.907} & 0.840 & 0.859 & 0.883 \\
& NWGI & \textbf{0.649} & 0.636 & 0.640 & 0.573 & 0.619 & 0.638 \\
\cline{2-8}
& Avg & 0.820 & 0.804 & \textbf{0.821} & 0.748 & 0.798 & 0.811 \\
\hline
\multirow{5}{*}{Multi-Task} & FPB & 0.861↓ & 0.845↓ & 0.870↓ & 0.766↓ & 0.836↓ & \textbf{0.873}↑ \\
& FiQA & 0.825↓ & \textbf{0.881}↑ & 0.863- & 0.737↓ & 0.822↓ & 0.870↑ \\
& TFNS & 0.890↓ & 0.880↓ & \textbf{0.892}↓ & 0.789↓ & 0.858↓ & 0.890↑ \\
& NWGI & 0.652↑ & 0.647↑ & 0.651↑ & 0.530↓ & 0.618↓ & \textbf{0.653}↑ \\
\cline{2-8}
& Avg & 0.807 & 0.813 & 0.819 & 0.701 & 0.784 & \textbf{0.822} \\
\hline
\multicolumn{2}{|c|}{Performance Gain} & -1.3\% & +0.7\% & -0.2\% & -4.7\% & -1.4\% & \textbf{+1.1\%} \\
\hline
\end{tabular}
\caption{Sentiment Analysis Instruction Tuning Results: The table reports detailed F1-scores for base models tuned during task-specific and multi-task phases on each sentiment analysis dataset. Arrows (↑↓) denote the influence of multi-task settings on Instruction Tuning results, with performance gains calculated between phases based on average F1 scores across all datasets.
}
\label{table:phase12_sa}
\vspace{-2mm}
\end{table}
\renewcommand{\arraystretch}{1.0}

\renewcommand{\arraystretch}{1.3}
\begin{table}[h]
\centering
\begin{tabular}{|c|c|c|c|c|c|c|c|}
\hline
Task & Phase & Llama2 & Falcon & MPT & BLOOM & ChatGLM2 & Qwen \\
\hline\hline
\multirow{3}{*}{NER} & Task-Specific & 0.637 & 0.619 & 0.615 & \textbf{0.729} & 0.645 & 0.679 \\
\cline{2-8}
& Multi-Task & 0.678↑ & 0.600↓ & 0.682↑ & \textbf{0.709}↓ & 0.629↓ & 0.666↓ \\
\cline{2-8}
& Performance Gain & +4.1\% & -1.9\% & \textbf{+6.7\%} & -2.0\% & -1.6\% & -1.3\% \\
\hline\hline
\multirow{3}{*}{HC} & Task-Specific & \textbf{0.942} & 0.940 & 0.938 & 0.930 & \textbf{0.942} & 0.936 \\
\cline{2-8}
& Multi-Task & \textbf{0.938}↓ & 0.932↓ & 0.928↓ & 0.898↓ & 0.932↓ & 0.922↓ \\
\cline{2-8}
& Performance Gain & \textbf{-0.4\%} & -0.8\% & -1.0\% & -3.2\% & -1.0\% & -1.4\% \\
\hline\hline
\multirow{3}{*}{RE} & Task-Specific & 0.395 & \textbf{0.428} & 0.309 & 0.425 & 0.340 & 0.371 \\
\cline{2-8}
& Multi-Task & 0.674↑ & 0.576↑ & 0.667↑ & \textbf{0.697}↑ & 0.557↑ & 0.640↑ \\
\cline{2-8}
& Performance Gain & +27.2\% & +14.8\% & \textbf{+35.8\%} & +27.2\% & +21.7\% & 26.9\% \\
\hline
\end{tabular}
\vspace{2mm}
\caption{Multi-Task Instruction Tuning Summary: The table reports entity-level F1 scores for NER, relation-only F1 for RE, and standard classification F1 for HC. It includes both task-specific and multi-task models for comparison. Arrows (↑↓) signify performance gains from multi-task settings, calculated in each task's last row.}
\label{table:phase12_ner}
\vspace{-2mm}
\end{table}
\renewcommand{\arraystretch}{1.0}

\subsection{Multi-Task Instruction Tuning}

\textbf{Performance Evaluation Setup.}
This subsection presents the performance evaluation of various LLMs following multi-task Instruction Tuning, and it provides a comparative analysis between models that underwent multi-task and task-specific tuning. Each model was trained on an aggregated dataset encompassing SA, HC, NER, RE, NER(CLS), and RE(CLS). Table \ref{table:phase12_sa} displays the results on the SA task and its sub-datasets, while Table \ref{table:phase12_ner} shows the results for NER, HC, and RE tasks.


\textbf{Performance in Classification Tasks.}
In classification tasks, such as SA and HC, most models displayed a minor decline in performance when concurrently trained with unrelated tasks. An exception to this trend, Qwen consistently enhanced its performance across all SA sub-datasets. Additionally, Falcon also showed improvement in specific areas. Conversely, BLOOM, which already had limited success during the task-specific tuning phase, suffered the most substantial performance degradation in classification tasks after multi-task tuning.

\textbf{Performance in Information Extraction Tasks.}
The scenario differs for information extraction tasks like NER and RE. For RE, all models exhibited significant improvement, likely due to the incorporation of RE(CLS) and additional financial NLP tasks, suggesting potential under-fitting during the task-specific tuning phase. While Llama2 and MPT exhibited progress in NER, not all models mirrored this improvement. Notably, these two models also displayed the most substantial enhancements in RE. BLOOM, in particular, made significant strides, outperforming Falcon and reaffirming its dominance in information extraction tasks. 

\textbf{Model Improvement Analysis.}
Despite its unsatisfying performance in the task-specific phase, MPT registered the most significant improvement in both NER and RE tasks, while Falcon and ChatGLM2 experienced moderate gains in both tasks, further underlining the interconnected performance dynamics in similar tasks.

\textbf{Summary of Findings.}
Llama2 and MPT not only displayed versatility by excelling in various tasks but also benefited from the multi-task learning environment. While models like Falcon, ChatGLM2, and Qwen maintained their baseline performances, BLOOM managed to enhance its strengths minimally. However, its weaknesses became more pronounced in a multi-task setting.

\renewcommand{\arraystretch}{1.3}
\begin{table}[h]
\centering
\begin{tabular}{|c|c|c|c|c|c|c|}
\hline
Dataset & Llama2 & Falcon & MPT & BLOOM & ChatGLM2 & Qwen \\
\hline
FPB & 0.621 & \textbf{0.791} & 0.599 & 0.576 & \textbf{0.803} & 0.576 \\
\hline
FiQA & 0.565 & \textbf{0.625} & 0.591 & 0.517 & \textbf{0.631} & 0.517 \\
\hline
\end{tabular}
\vspace{2mm}
\caption{Zero-shot Sentiment Analysis Results: The zero-shot F1-scores on the Sentiment Analysis test datasets are reported for all base models. The two best models are highlighted in bold.}
\label{table:phase3_sa}
\vspace{-2mm}
\end{table}
\renewcommand{\arraystretch}{1.0}

\subsection{Instruction Tuning for Zero-shot Ability}

\textbf{Training Setup and Modification.}
This subsection delineates the results of our examination on the zero-shot abilities of LLMs post-Instruction Tuning. For this evaluation, the LLMs were trained on three distinct classification tasks: HC, NER(CLS), and RE(CLS). Initially, we endeavored to employ the original NER and RE tasks for training purposes. However, these did not sufficiently activate the models' zero-shot abilities due to insufficient instruction diversity. This limitation led to issues such as model hallucination and inconsistency in response to the provided options. 

\textbf{Task Reformulation to Mitigate Hallucination.}
To mitigate this, we reformulated NER and RE tasks into classification tasks, thereby narrowing the task gap with the target SA task. Furthermore, considering the challenge in distinguishing between neutral sentiments and positive/negative ones in a zero-shot setting, all samples labeled "neutral" were excluded from the study. 

\textbf{Comparative Performance Insights.}
Table \ref{table:phase3_sa} presents the mean F1-score results for FiQA and FPB. ChatGLM2 stood out in zero-shot tasks, likely due to its chat-centric tuning. Falcon followed closely in overall performance. Llama2 and MPT, while lagging behind ChatGLM2 and Falcon, showed promise in classifying a subset of the samples. However, both BLOOM and Qwen struggled, often misclassifying responses as "positive". BLOOM's results were expected given its past struggles with classification, but Qwen's suboptimal performance was surprising.

\textbf{Generalization Capability Insights.}
These findings are particularly illuminating as neither ChatGLM2 nor Falcon excelled in the task-specific and multi-task Instruction Tuning phases but demonstrated significant generalization capabilities, understanding unseen instructions, and making accurate classifications during the zero-shot tasks.

\section{Conclusion and Future Work}

In conclusion, this paper presented an Instruction Tuning paradigm that includes task-specific, multi-task, and zero-shot instruction tuning of LLMs within the financial sector. Our work articulated and showcased the diverse capabilities and potential limitations of different LLMs when subjected to various NLP tasks integral to finance. Through rigorous experimentation and analysis, the paper unveiled distinct performance patterns and offered valuable insights into how these models can be efficiently and effectively employed for specific financial applications.

Future work will focus on integrating additional open-source base models, investigating larger models with parameter sizes between 13 and 100 billion, and deepening efforts to enhance the robustness and generalization capabilities of Large Language Models (LLMs). We'll also develop strategies to reduce task interference and hallucination \cite{ji2023survey}, ensuring accurate and reliable model responses across diverse tasks.

\medskip
\small
\bibliographystyle{plain}
\bibliography{ref}

\end{document}